\documentclass[wcp]{jmlr}

\usepackage{longtable}
\usepackage{booktabs}
\usepackage{algorithmic}
\usepackage{lineno}

\makeatletter
\let\Ginclude@graphics\@org@Ginclude@graphics
\makeatother

\title[CMP-IRRT*: Height-Adaptive Planning]{CMP-IRRT*: A Perception-Assisted Height-Adaptive Planner for Quadruped Robots\textsuperscript{\textdagger}}

\author{
\Name{Mingfan Zhao} \Email{zhaomf7@mail2.sysu.edu.cn}\\
\Name{Wendong Mao}\textsuperscript{*} \Email{maowd@mail.sysu.edu.cn}\\
\Name{Zhongfeng Wang}\textsuperscript{*} \Email{wangzf83@mail.sysu.edu.cn}\\
\addr School of Integrated Circuits, Shenzhen Campus of Sun Yat-sen University, Shenzhen 518107, China\\
\textsuperscript{*}Corresponding authors\\
\textsuperscript{\textdagger}Accepted at the 18th Asian Conference on Machine Learning (ACML 2026).
}

\begin{document}

\maketitle
\thispagestyle{plain}
\pagestyle{plain}

\begin{abstract}
Quadruped robots can traverse low obstacles, but many 2D planning pipelines still model obstacles as binary occupied regions and rely on sampling-based search that can be inefficient under a limited budget. We propose a perception-assisted height-adaptive planning framework based on CMP-IRRT*, a Channel Mamba PointNet-guided Informed RRT* planner. Given a calibrated top-view RGB observation, the perception module estimates obstacle regions and converts depth predictions into a ground-relative height map. The planner then performs height-conditioned collision checking, treating high obstacles as blocked while allowing low obstacles to be traversed, and uses the CMP guide to bias sampling toward promising regions while retaining standard free-space and informed sampling fallbacks. Experiments on 2D planning benchmarks show that CMP-IRRT* reduces explored nodes and iterations compared with classical and neural-guided baselines, and a controlled ablation supports the contribution of the Mamba-based guide. In constructed traversability-aware scenarios, the proposed planner reduces path length by up to 16.3\% when low obstacles are traversable, and a Unitree Go2 demonstration further shows executable bypassing and traversal behaviors. Our code is publicly available at {\hypersetup{urlcolor=black}\url{https://github.com/MingfanZhao/height-adaptive-planner}}.
\end{abstract}

\begin{keywords}
\leftskip 19pt\rightskip 19pt
Path planning; Height-adaptive planning; Guided sampling; Quadruped robots
\end{keywords}

\section{Introduction}

Quadruped robots can move in environments that are difficult for wheeled platforms, partly because their legs allow them to step over or traverse low obstacles. However, many 2D path planning pipelines still model the environment as a binary occupancy map: an obstacle is either free or blocked. This representation is simple and widely used, but it can lead to unnecessary detours when an obstacle is low enough for a legged robot to cross safely. For quadruped navigation, the planner should therefore consider not only obstacle position, but also obstacle height and the robot's traversal capability.

Sampling-based planners such as RRT* and Informed RRT* are attractive for robot path planning because they can handle complex environments and improve path quality over time. Learning-guided planners further improve finite-budget search by predicting promising regions of the state space. Nevertheless, these methods usually focus on where to sample and how to avoid collision, while treating all detected obstacles as non-traversable. In this work, we study a complementary question: how can visual perception and robot-specific traversability constraints be incorporated into a planning pipeline so that low obstacles can be crossed instead of always bypassed?

We propose a visual perception-assisted traversability-aware path planning pipeline for quadruped robots. The system uses a calibrated top-view RGB observation to estimate obstacle shape, location, and height. These obstacle attributes are combined with a robot-specific height threshold to decide whether an obstacle should be treated as blocked or traversable during planning. The resulting height-conditioned collision check is integrated with a Channel Mamba PointNet-guided Informed RRT* planner (CMP-IRRT*) to improve search efficiency under a limited planning budget. The output path is decoded into velocity commands and motion modes for execution on a quadruped robot.

This paper does not claim a new optimality theory for RRT-based planning. Instead, it contributes a practical formulation and implementation of perception-conditioned traversability-aware planning. The main contributions are:
\begin{itemize}
\item We formulate a traversability-aware planning problem for quadruped robots, where obstacle handling depends on both perceived obstacle height and robot crossing capability.
\item We design a modular perception-assisted planning pipeline that estimates obstacle geometry from a top-view RGB image and uses height-conditioned collision checking to decide whether to bypass or traverse obstacles.
\item We introduce a Channel Mamba PointNet-guided IRRT* planner to improve finite-budget sampling efficiency by predicting promising search regions.
\item We validate the framework on 2D planning benchmarks, constructed traversability-aware scenarios, and a Unitree Go2 demonstration, showing reduced explored nodes and iterations for CMP-IRRT* and up to a 16.3\% path-length reduction when low obstacles are traversable.
\end{itemize}

\section{Related Work}

\subsection{Sampling-Based and Learning-Guided Planning}
Classical path planning methods include graph-search algorithms such as A*~\citep{A_star_1968}, sampling-based planners such as RRT*~\citep{RRT_star_2011}, and potential-field methods~\citep{APF_1985}. Informed RRT* improves the efficiency of RRT* by restricting sampling to an ellipsoidal subset once an initial solution is found~\citep{Informed_RRT_star}. Learning-guided planners further use neural networks to predict useful search regions or guidance maps. Neural A* encodes planning instances into a differentiable search guidance map~\citep{Neural_A_star}, while NIRRT* uses point-cloud constraints to guide Informed RRT*~\citep{NIRRT_Star}. These approaches improve search efficiency, but they usually treat obstacles as binary collision regions. Our work is complementary: it introduces a traversability-aware collision model for quadruped robots and combines it with a learned sampling guide.

\subsection{Traversability-Aware Planning for Legged Robots}
Legged robots differ from wheeled robots because they can step over low obstacles and exploit multiple locomotion modes. Reviews of legged robot motion and design emphasize that locomotion capability depends on the robot's morphology and control constraints~\citep{Legged_Robot_review}. In many planning pipelines, however, this capability is not explicitly reflected in the 2D map representation. We focus on a practical setting in which obstacle height is estimated from visual perception and compared with a robot-specific crossing threshold. This allows the planner to treat some obstacles as traversable while still avoiding obstacles that exceed the robot's capability.

Representative terrain-referenced elevation-mapping methods maintain 2.5D per-cell terrain geometry from range sensing and state estimation for legged locomotion and navigation~\citep{Fankhauser2014Elevation,Miki2022Elevation}. Our method addresses a complementary operating regime. It converts a calibrated top-view RGB observation into an image-aligned ground-relative height map and directly supplies object-level height information to CMP-IRRT*, without maintaining an online terrain-map fusion pipeline. Elevation maps retain dense terrain detail, but they depend on range measurements, pose estimates, and map fusion. Our formulation instead emphasizes direct perception-to-planning integration for the approximately planar-topped obstacles in the static scenes evaluated here. It remains sensitive to monocular depth errors, calibration, occlusion, and non-planar objects. A rigorous cross-paradigm comparison would require common sensing inputs and mapping conditions. We therefore do not directly compare performance with elevation-map planners in the present study. Our evaluation instead focuses on finite-budget sampling and height-conditioned traversability within the calibrated top-view setting studied here.

\subsection{Vision-Based Obstacle Perception}
Vision-based perception provides the geometric information required by traversability-aware planning. Object detection models such as YOLO~\citep{YOLO,YOLOv11} can identify obstacle regions, while monocular depth models such as Depth Anything V2~\citep{Depth_Anything_V2} provide depth estimates from RGB images. In this paper, we do not propose a new depth estimation model. Instead, we use existing perception modules to provide obstacle position and height estimates for the planner. This design keeps the planning contribution separate from the perception backbone and makes the assumptions of the pipeline explicit.

\begin{figure*}[t]
\centering
\includegraphics[width=0.98\textwidth]{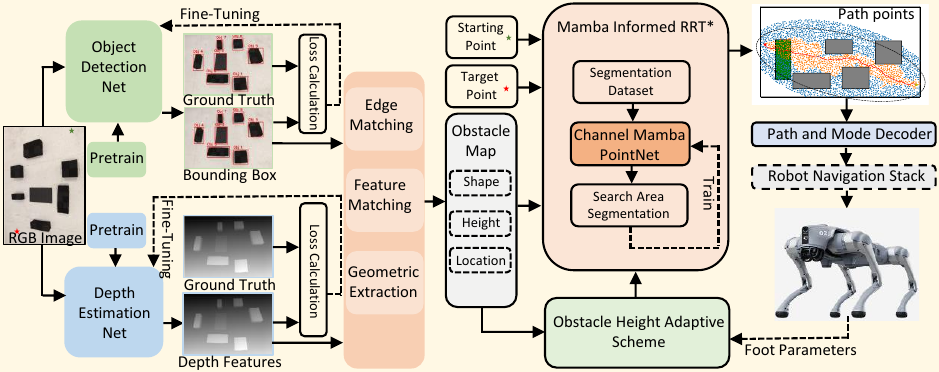} 
\caption{Overview of the proposed perception-assisted traversability-aware planning pipeline. The perception module estimates obstacle geometry from a calibrated top-view RGB observation. The planning module uses height-conditioned collision checking and CMP-guided informed sampling to find a low-cost path under a finite planning budget. The execution module decodes the path and traversal mode into robot commands.}
\label{fig_framework}

\end{figure*}

\section{Method}

\subsection{Problem Setup and Assumptions}
Figure~\ref{fig_framework} illustrates the proposed planning pipeline. We consider a quadruped robot navigating in a static or slowly changing environment represented by a 2D planning map. The input is a calibrated top-view RGB observation, together with start and goal positions. The camera may be an external overhead camera or another calibrated sensing setup that provides a top-view image. The image-to-map scale is assumed to be known from calibration. The perception module estimates obstacle regions and approximate obstacle heights; these estimates are then used by the planner rather than optimized jointly with it.

This setup is intentionally modular. The method is not an end-to-end learned controller in the strict machine learning sense. Instead, it connects perception, planning, and command decoding so that the planner can make height-aware decisions. The current implementation plans before execution and does not perform closed-loop online replanning during robot motion.

\subsection{Perception-Assisted Planning Pipeline}
The pipeline has three modules. The perception module applies object detection and depth estimation to the top-view RGB image to estimate obstacle shape, location, and height. The planning module combines these obstacle attributes with the robot's traversable height threshold and runs CMP-IRRT* to find a low-cost path. The execution module decodes the path and the selected traversal mode into robot velocity commands.

Unlike a binary occupancy planner, our planner distinguishes between high obstacles that should be avoided and low obstacles that may be crossed. This distinction is important for quadruped robots because a safe crossing can reduce unnecessary detours. The output of the planner is therefore not only a geometric path, but also a sequence of mode decisions indicating whether the robot should bypass or traverse encountered obstacles.

\subsection{Traversability-Aware Obstacle Handling}
In a 2D planning map, conventional planners usually bypass all occupied regions. For quadruped robots, this can be overly conservative. We define a robot-specific height threshold $H_{\alpha}$ and use it to decide whether an obstacle is traversable. The threshold depends on robot morphology and locomotion constraints:
\begin{equation}
H_{\alpha} = \min \{ P_{foot} \times \max \{ \cos \theta \}, H_{round} \}.
\end{equation}
Here $P_{foot}$ is the maximum foot clearance, $\theta$ is the thigh-joint angle, $\max\{\cos\theta\}$ is evaluated over its feasible range for traversal, and $H_{round}$ is the robot-specific maximum obstacle-crossing height.

The perception module first estimates a depth map and converts it into a ground-relative height map using the calibrated top-view geometry:
\begin{equation}
I_{\mathrm{height}} = \mathcal{T}_{\mathrm{calib}}(I_{\mathrm{depth}}, \Pi_{\mathrm{ground}}),
\end{equation}
where $\Pi_{\mathrm{ground}}$ denotes the ground-plane reference and $\mathcal{T}_{\mathrm{calib}}$ denotes the calibration-based depth-to-height transformation. The obstacle height estimates are then given by:
\begin{equation}
H_N = \{ h_1,h_2,...,h_N \},
\end{equation}
\begin{equation}
h_i = I_{\mathrm{height}}[x_i,y_i],
\end{equation}
\begin{equation}
\{x_i, y_i \}= \{x_{lt}+W_i/2,y_{lt}+L_i/2\}.
\end{equation}
Here $N$ is the number of obstacles, $I_{\mathrm{depth}}$ is the estimated depth map, $I_{\mathrm{height}}$ is the calibrated ground-relative height map, and $L_i$ and $W_i$ are the estimated obstacle length and width. These estimates are approximate; the planner uses them to select a traversability mode rather than to guarantee precise 3D reconstruction.

The current implementation samples one height value at the center of each detected bounding box. For the approximately rectangular, planar-topped obstacles used in the controlled experiments, this point lies away from the box boundary and provides a representative top-surface estimate. It is not a robust estimator for arbitrary non-planar, irregular, or occluded objects. Region-based height estimates and uncertainty margins are possible extensions, but they are not used in the reported experiments.

We incorporate the traversability decision into collision checking. Obstacles with estimated height above $H_{\alpha}$ are treated as blocked, while obstacles below the threshold can be crossed by selecting the traversal mode. Algorithm~\ref{alg:cmirrt} summarizes the resulting planner.

\subsection{Channel Mamba PointNet-Guided Informed RRT*}
The learned guide is used to improve sampling efficiency under a finite planning budget. RRT* can approach an optimal solution asymptotically, but its convergence can be slow. Informed RRT* improves sampling once an initial solution is available, and NIRRT* further uses neural guidance to focus sampling on promising regions. We adopt this learning-guided planning perspective and use a Channel Mamba PointNet (CMP) guide to predict a compact search region.

\begin{algorithm}[!t]
\caption{CMP-IRRT* with Height-Conditioned Collision Checking}
\label{alg:cmirrt}
\begin{algorithmic}[1]
\STATE Initialize tree $\mathcal{T}$ with $x_{\mathrm{start}}$
\STATE $c^{\mathrm{best}} \leftarrow \infty$, $c^{\mathrm{update}} \leftarrow \infty$
\STATE Set guide sampling probability $p_g$
\STATE $X_{\mathrm{guide}} \leftarrow \text{ChannelMambaPointNetGuide}(X_{\mathrm{free}}, x_{\mathrm{start}}, x_{\mathrm{goal}})$
\STATE Set robot traversability threshold $H_\alpha$
\WHILE{not converged}
    \IF{$\text{Rand}() < p_g$}
        \STATE $x_{\mathrm{rand}} \leftarrow \text{SampleFromGuide}(X_{\mathrm{guide}})$
    \ELSIF{$c^{\mathrm{best}} < \infty$}
        \STATE $x_{\mathrm{rand}} \leftarrow \text{SampleFromEllipsoid}(x_{\mathrm{start}}, x_{\mathrm{goal}}, c^{\mathrm{best}})$
    \ELSE
        \STATE $x_{\mathrm{rand}} \leftarrow \text{SampleFree}()$
    \ENDIF
    \STATE $x_{\mathrm{near}} \leftarrow \text{NearestNeighbor}(\mathcal{T}, x_{\mathrm{rand}})$
    \STATE $x_{\mathrm{new}} \leftarrow \text{Steer}(x_{\mathrm{near}}, x_{\mathrm{rand}})$
    \STATE \textit{// Apply height-conditioned collision checking}
    \IF{$\text{HeightAwareCollisionFree}(x_{\mathrm{near}}, x_{\mathrm{new}}, H_{\alpha})$}
        \STATE $\mathcal{T}.\text{add}(x_{\mathrm{new}})$
        \STATE $\text{Rewire}(\mathcal{T}, x_{\mathrm{new}})$
    \ENDIF
    \IF{$c^{\mathrm{best}} < \alpha \cdot c^{\mathrm{update}}$}
        \STATE $X_{\mathrm{focus}} \leftarrow \{x \mid \|x - x_{\mathrm{start}}\| + \|x - x_{\mathrm{goal}}\| \leq c^{\mathrm{best}}\}$
        \STATE $X_{\mathrm{guide}} \leftarrow \text{ChannelMambaPointNetGuide}(X_{\mathrm{focus}} \cap X_{\mathrm{free}})$
        \STATE $c^{\mathrm{update}} \leftarrow c^{\mathrm{best}}$
    \ENDIF
    \IF{$\neg\text{IsConnected}(X_{\mathrm{guide}})$}
        \STATE $X_{\mathrm{guide}} \leftarrow \text{NeuralConnect}(X_{\mathrm{guide}})$
    \ENDIF
\ENDWHILE
\end{algorithmic}
\end{algorithm}

NeuralConnect constructs a radius graph over the start, goal, and accumulated predicted guide points and checks start--goal connectivity using breadth-first search (BFS). If disconnected, it selects a frontier of the start-reachable component, updates the local start and goal masks, unions the next prediction with the accumulated guide, and repeats for at most five classifier trials.

The guide is used as a probabilistic sampling bias: with probability $p_g$, the planner samples from the CMP-predicted guide set, and otherwise falls back to standard free-space or informed sampling.

The motivation for using Mamba is the need to capture long-range connectivity in point-cloud representations of planning maps. A useful guide should not only recognize local obstacle geometry, but also infer corridor-level structure between start and goal. Mamba models long-range dependencies with linear-time sequence processing~\citep{mamba,SSMs,Vision_Mamba}, while channel attention helps emphasize informative feature channels~\citep{CBAM}. We therefore combine Mamba and channel attention with a PointNet++-style backbone to improve guide quality for planning.

Valid free-space candidates are sampled uniformly. The initial guide input targets 2048 points and uses farthest-point downsampling when more candidates remain after filtering; a focused update may contain fewer points. We retain the returned array order and apply no learned, raster, or coordinate sorting. For a point set of size $N$, PointNet++ produces 128-dimensional features, which are transposed to a $B\times N\times128$ sequence, processed by six Mamba blocks, and fused with the channel-attention-weighted PointNet++ branch. We therefore do not claim strict permutation invariance for the Mamba stage. All controlled guide variants receive the same initial sampled point cloud under matched seeds. Table~\ref{tab:mamba_ablation_2d} is an architecture ablation rather than an ordering-sensitivity test.

\begin{figure}[!t]
\vspace{-8pt}
\centering
\setlength{\abovecaptionskip}{3pt}
\setlength{\belowcaptionskip}{-4pt}
\includegraphics[width=0.98\columnwidth,trim=0 2pt 0 2pt,clip]{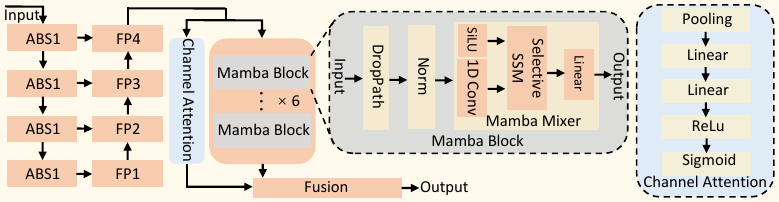}
\caption{Structure of the Channel Mamba PointNet guide used to predict promising search regions for CMP-IRRT*.}
\label{mamba_pointnet}
\vspace{-8pt}
\end{figure}

Our CMP module is used as a planning guide rather than as a standalone planner. It proposes a focused subset of the free space, and the sampling-based planner still performs collision checking, tree expansion, and rewiring. This keeps the planning process compatible with robot-specific constraints such as the traversability threshold.

\subsection{Path and Mode Decoder}
After planning, the path is decoded into velocity commands and traversal modes. The path is represented as
\begin{align}
\text{Path\_Points} &= \{\text{point}_1, \text{point}_2, \ldots, \text{point}_K\}, \\
\text{point}_i &= \{x_i, y_i\}, \\
v_{i_x} &= \frac{x_{i+1} - x_i}{T_{\text{scale}}}, \\
v_{i_y} &= \frac{y_{i+1} - y_i}{T_{\text{scale}}}, \\
\text{yaw}_i &= \operatorname{atan2}\left(y_{i+1} - y_i,\;x_{i+1} - x_i\right), \\
v_{i_{\text{yaw}}} &= \frac{\text{yaw}_{i+1} - \text{yaw}_i}{T_{\text{scale}}}, \\
mode &=
\begin{cases}
A & h_i \geq H_{\alpha} \\
B & h_i < H_{\alpha}
\end{cases}.
\end{align}
Here $x$ and $y$ are path coordinates, $T_{\text{scale}}$ is the interval between motion commands, $yaw$ is the robot heading, and $mode$ determines whether the robot bypasses or traverses an obstacle.

\section{Experiments}

We evaluate the proposed pipeline in two parts. First, we test the CMP-IRRT* planner on standard 2D planning problems to examine finite-budget search efficiency. Second, we evaluate the traversability-aware pipeline in constructed environments with obstacles of different heights and demonstrate deployment on a Unitree Go2 robot. The experiments are intended to show practical feasibility and average performance in the evaluated scenarios, rather than to establish a new theoretical optimality result.

All model training and server-based evaluation experiments reported in this paper were conducted on servers equipped with NVIDIA GeForce RTX 4090 GPUs using Python 3.9.16 and PyTorch 1.13.1.

\begin{figure*}[t]
\centering
\includegraphics[width=0.98\textwidth]{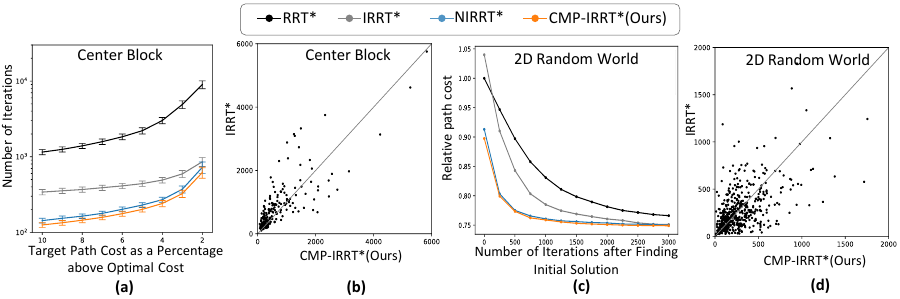}
\caption{Comparison of planning performance on Center Block and 2D Random World problems. The compared methods include RRT*~\citep{RRT_star_2011}, IRRT*~\citep{Informed_RRT_star}, and NIRRT*~\citep{NIRRT_Star}. The plots report finite-budget planning behavior, including iterations to reach a cost threshold and path-cost improvement after an initial solution is found.}
\label{fig_mirrt_star_comparison}
\end{figure*}

\subsection{CMP-IRRT* Simulation Experiments}
\subsubsection{Experimental Settings}
We used 1000 random worlds as the training dataset and 150 as the test dataset. For each random world, we run the A* algorithm in pixel space to obtain a reference path. The random worlds are represented as 3D point clouds, with the $z$ coordinate set to 0 for the 2D map. The reference path is used as supervision for learning a guide region. Specifically, the area within an $\eta$-radius of each reference path point is treated as part of the target guide region, forming an approximately continuous strip.

To evaluate the learned sampling guide itself, we remove the height-aware traversability component in this experiment and compare CMP-IRRT* with existing planners on general 2D planning datasets. Following the setting of NIRRT*~\citep{NIRRT_Star}, we use Center Block and 2D Random World problems. Center Block evaluates whether a planner can focus sampling in a large free space, while Random World evaluates generality across varied obstacle layouts.

\subsubsection{Metrics and Results}
For Center Block, we measure the number of iterations required to reach a path-cost threshold. For Random World, we focus on initial-solution search efficiency and post-initial path optimization. Specifically, we report the following metrics:
\begin{samepage}
\begin{itemize}
\setlength{\itemsep}{0pt}
\setlength{\parsep}{0pt}
\setlength{\topsep}{2pt}
\item Initial Time: the planner-level elapsed time to the first feasible path, including guide construction and any NeuralConnect calls performed before that path is found.
\item Nodes: the average number of nodes explored when the initial path is found.
\item Iterations: the average number of iterations when the initial path is found.
\item Final Cost: the average path length recorded after 18 additional seconds of post-initial optimization.
\end{itemize}
\end{samepage}
These metrics emphasize finite-budget planning performance, which is the intended use case of the CMP guide.

Figure~\ref{fig_mirrt_star_comparison} shows that CMP-IRRT* generally reaches lower-cost solutions with fewer iterations in the evaluated environments. The results indicate that the learned guide can focus sampling on useful regions of the state space. We report mean values and avoid claiming statistical significance beyond the evaluated test set.

\begin{table}[t]
\centering
\caption{Average comparison results on the 2D Random World dataset. Lower values are better for all metrics.}
\label{tab:time_comparison_1}
\begingroup
\small
\setlength{\tabcolsep}{4pt}
\begin{tabular}{lcccc}
\toprule
Method & RRT* & IRRT* & NIRRT* & CMP-IRRT*(Ours)  \\
\midrule
Initial Time(s) & \textbf{0.17}   & 0.19   & 1.81   &  1.37 \\
Nodes & 241.86   & 251.61   & 102.92   & \textbf{99.49}   \\
Iterations & 299.04   & 312.75   & 151.69   &  \textbf{146.55}  \\
Final Cost & 233.01   & 232.63   & 206.51   & \textbf{203.41}   \\
\bottomrule
\end{tabular}
\endgroup
\end{table}

Table~\ref{tab:time_comparison_1} summarizes the average Random World results. Traditional RRT* and IRRT* find initial paths quickly because they do not require neural inference. Among the learning-guided methods, CMP-IRRT* has a lower observed mean initial-solution planning time than NIRRT* in this setting (1.37 s vs. 1.81 s), while also obtaining a lower observed mean final cost. These results support the use of CMP as a finite-budget sampling guide, although broader statistical testing would be needed for stronger claims.

\subsubsection{Ablation on Guide Architecture}
We further conduct a normalized controlled rerun to isolate the effect of the guide architecture on initial-solution search efficiency. This ablation uses fixed guide variants and fixed random seeds, and reports only normalized search-efficiency ratios for within-table comparison. Unlike the full planner comparison in Table~\ref{tab:time_comparison_1}, it is not intended to reproduce or match the absolute values across all metrics. We use the PointNet++~\citep{PointNetPP_2017} guide as the baseline and normalize the average number of nodes and iterations required to obtain an initial feasible path.

\begin{table}[t]
\centering
\caption{Normalized controlled ablation of the guide architecture on 2D Random World. Nodes Ratio and Iterations Ratio denote the average number of nodes and iterations required to find the initial feasible path, normalized by the PointNet++ guide baseline. Lower values indicate better initial-solution search efficiency.}
\label{tab:mamba_ablation_2d}
\begingroup
\footnotesize
\setlength{\tabcolsep}{2pt}
\begin{tabular}{lcccc}
\toprule
Guide & Mamba & Channel Attn. & Nodes Ratio & Iterations Ratio \\
\midrule
PointNet++ & No & No & 1.000 & 1.000 \\
PointNet++ and Mamba & Yes & No & 0.984 & 0.980 \\
CMP & Yes & Yes & \textbf{0.919} & \textbf{0.920} \\
\bottomrule
\end{tabular}
\endgroup
\end{table}

Table~\ref{tab:mamba_ablation_2d} shows that adding Mamba reduces the normalized node and iteration ratios to 0.984 and 0.980, respectively. The full CMP guide further reduces them to 0.919 and 0.920, corresponding to about 8.1\% fewer nodes and 8.0\% fewer iterations than the PointNet++ guide. These results suggest that the Mamba-based representation and channel-attention fusion help focus sampling on more useful regions of the state space.

\FloatBarrier
\subsubsection{Paired Repeated Evaluation}
We additionally evaluated one protocol-matched NIRRT*/CMP-IRRT* model pair (training seed 100) on the 150 Random World test maps, using one start--goal pair and three paired planner seeds per map. Both methods used the same training protocol, maps, start--goal pairs, planner seeds, evaluation settings, and A*-generated offline supervision; neither method received A* information at test time. Each method was evaluated in 450 runs under each setting, giving 1800 method-runs in total. Success rate is computed at the run level, whereas the three planner seeds are first summarized for each map before continuous metrics are reported as mean $\pm$ standard deviation over 150 maps.

\begin{table}[h]
\centering
\caption{Paired repeated results on the 150-map Random World test set. Continuous metrics are reported as map-level mean and standard deviation, while success rates are computed over individual runs. Path cost is measured in map units. Lower is better except for success rate.}
\label{tab:paired_repeated_2d}
\begingroup
\setlength{\tabcolsep}{4pt}
\begin{tabular}{p{0.58\linewidth}cc}
\toprule
Evaluation setting and metric & NIRRT* & CMP-IRRT* \\
\midrule
Table~\ref{tab:time_comparison_1} protocol: success rate & 100.0\% & 100.0\% \\
Table~\ref{tab:time_comparison_1} protocol: initial-solution time (s) & $1.813 \pm 0.770$ & $1.380 \pm 0.650$ \\
Table~\ref{tab:time_comparison_1} protocol: final path cost after 18 s of post-initial optimization & $207.27 \pm 26.79$ & $204.11 \pm 25.73$ \\
2000-expansion budget: success rate & 99.8\% & 100.0\% \\
2000-expansion budget: iterations to first solution & $151.71 \pm 106.24$ & $146.61 \pm 99.87$ \\
2000-expansion budget: terminal path cost & $207.06 \pm 26.62$ & $203.87 \pm 25.29$ \\
\bottomrule
\end{tabular}
\endgroup
\end{table}

Under the 2000-expansion budget, one of the 450 NIRRT* runs did not return a path; success rate is therefore reported separately from the continuous metrics. Table~\ref{tab:paired_repeated_2d} is an additional paired repeated evaluation and does not replace the absolute comparison in Table~\ref{tab:time_comparison_1} or the normalized architecture ablation in Table~\ref{tab:mamba_ablation_2d}. Under the Table~\ref{tab:time_comparison_1} protocol, both methods achieved the same observed success rate, while CMP-IRRT* had lower observed mean initial-solution time and final path cost. Under the equal-expansion setting, CMP-IRRT* also had lower observed mean first-solution iterations and terminal path cost. These results are descriptive evidence for one training pair and characterize variation across maps and planner seeds rather than across training seeds. The paired runs in Table~\ref{tab:paired_repeated_2d} followed the hardware and software setting reported above.

\FloatBarrier
\subsection{Traversability-Aware Planning}
Our depth estimation module uses a pre-trained Depth Anything V2~\citep{Depth_Anything_V2} model and is fine-tuned for the target setting. The object detection module uses a YOLOV11 \citep{YOLOv11} model from the YOLO~\citep{YOLO} series. These perception modules provide obstacle regions and approximate height estimates to the planner.

\begin{figure}[t]
\centering
\includegraphics[width=0.98\columnwidth]{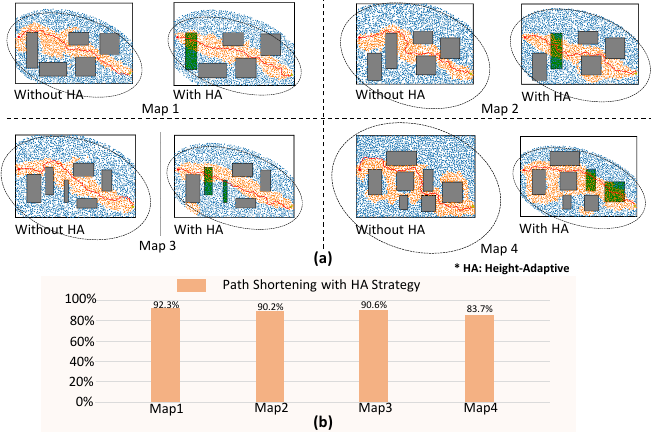}
\caption{Traversability-aware planning results. (a) Visualization on four maps with low-height obstacles that can be traversed. Green obstacles denote obstacles below the threshold $H_{\alpha}$. (b) Path shortening after applying height-conditioned traversal decisions.}
\label{results_mirrt_star}
\end{figure}

We construct maps containing multiple obstacles with different heights and use the perception-assisted pipeline to plan paths. Figure~\ref{results_mirrt_star}(a) shows examples where low-height obstacles are treated as traversable, while higher obstacles are avoided. To evaluate the effect of height-aware traversal, we compare the same CMP-guided planner with and without height-conditioned obstacle handling on four constructed maps. The number of iterations is set to 1500, and the reported values are averaged over 10 planning runs. Figure~\ref{results_mirrt_star}(b) shows that enabling height-conditioned traversability reduces the mean planned path length. In the evaluated setting, with at most two traversable obstacles per map, the reductions range from 7.7\% to 16.3\%. These reductions are attributed to height-conditioned traversability within CMP-IRRT*, rather than to the CMP guide alone.

\subsection{Controlled Real-Robot Evaluation}
\begin{figure}[!t]
\centering
\includegraphics[width=0.7\columnwidth]{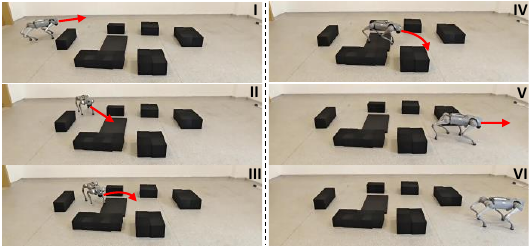}
\caption{Real-robot demonstration on a Unitree Go2 robot. The robot executes planned motions that include both obstacle bypassing and traversal of low obstacles.}
\label{real_world_move}
\end{figure}

We evaluated three configurations of the same CMP-IRRT* planner in a fixed mixed-obstacle scene, while keeping the CMP weights and planner settings unchanged. Binary-CMP treated every detected obstacle as blocked, Oracle-height CMP used measured obstacle heights, and Full predicted-height CMP used heights estimated by the monocular perception pipeline.

\begin{table}[!htbp]
\centering
\caption{Controlled Unitree Go2 trials. Task completion uses all trials for each configuration, whereas executed-path length is reported as mean and standard deviation on the common matched subset of ten trials.}
\label{tab:go2_repeated_trials}
\begingroup
\setlength{\tabcolsep}{3pt}
\begin{tabular}{p{0.21\linewidth}p{0.15\linewidth}p{0.22\linewidth}p{0.31\linewidth}}
\toprule
Configuration & \shortstack[l]{Task\\completion} & \shortstack[l]{Executed path,\\matched n=10} & Intervention/safety \\
\midrule
Binary-CMP & 10/10 & $8.56 \pm 0.11$ m & 0 interventions \\
\shortstack[l]{Oracle-height\\CMP} & 10/10 & $7.25 \pm 0.09$ m & 0 interventions \\
\shortstack[l]{Full predicted-\\height CMP} & 18/20 & $7.71 \pm 0.13$ m & \shortstack[l]{2/20 trials required\\intervention; 0 falls;\\0 body collisions} \\
\bottomrule
\end{tabular}
\endgroup
\end{table}

On the common matched subset of ten trials, Oracle-height CMP and Full predicted-height CMP reduced mean executed-path length relative to Binary-CMP by 15.3\% and 9.9\%, respectively. Across all Full predicted-height CMP trials, 18 of 20 tasks were completed, while two trials required intervention. No fall or body collision was observed.

In separate threshold tests, the robot completed 10/10 traversals at $0.7H_{\alpha}$ and 9/10 at $0.9H_{\alpha}$; the latter included one intervention, with no unsafe contact or fall. The height decision was correct in 18/20 traversable-height cases and 20/20 blocked-height cases. Both errors were conservative false-blocked decisions, and no false-traversable decision was observed. The Full predicted-height configuration avoided the high obstacle in 20/20 encounters.

These evaluations use their stated denominators and are not pooled into a single trial count. They were conducted with a calibrated top-view camera, approximately planar-topped obstacles, a static scene, and planning before execution. The results provide controlled physical evidence under these conditions rather than a broad-terrain evaluation or a general safety guarantee. Figure~\ref{real_world_move} shows representative moments during execution, including traversal and bypassing behaviors.

\section{Limitations}

The current pipeline has several limitations. First, it assumes a calibrated top-view observation and a static or slowly changing environment during planning. It therefore does not directly solve onboard first-person perception or online replanning in dynamic scenes. Second, height estimation from monocular RGB images can be noisy, and incorrect height estimates may lead to incorrect traversal decisions. Third, the CMP guide is evaluated as a finite-budget sampling aid; the current results do not establish statistical significance across a large benchmark suite. Finally, the method uses a modular perception-planning-control pipeline rather than an end-to-end learned controller.

\section{Conclusion}

We presented CMP-IRRT*, a perception-assisted height-adaptive path planning framework for quadruped robots. The method combines calibrated top-view perception, ground-relative height estimation, height-conditioned collision checking, and CMP-guided informed sampling so that the planner can both bias search toward promising regions and decide whether low obstacles should be traversed or high obstacles should be bypassed. Experiments on 2D planning benchmarks show improved finite-budget search behavior in the evaluated settings, while constructed traversability-aware scenarios show path shortening when low obstacles are treated as traversable. A Unitree Go2 demonstration further illustrates that the planned paths and traversal modes can be converted into executable robot commands. These results suggest that coupling learned sampling guidance with height-adaptive obstacle handling is a practical direction for reducing unnecessary detours in quadruped path planning.

\acks{This work was supported in part by the National Natural Science Foundation of China under Grant 62504260, in part by the Jiangsu Provincial Science and Technology Major Special Project BG2024032, and in part by Shenzhen Science and Technology Program under Grants ZDCY20250901112302003 and ZDCY20250901112804006.}

\bibliography{reference_daoa}

\end{document}